# Incremental Boosting Convolutional Neural Network for Facial Action Unit Recognition

**Shizhong Han, Zibo Meng, Ahmed Shehab Khan, Yan Tong**
Department of Computer Science & Engineering, University of South Carolina, Columbia, SC
{han38, mengz, akhan}@email.sc.edu, tongy@cse.sc.edu

## Abstract

Recognizing facial action units (AUs) from spontaneous facial expressions is still a challenging problem. Most recently, CNNs have shown promise on facial AU recognition. However, the learned CNNs are often overfitted and do not generalize well to unseen subjects due to limited AU-coded training images. We proposed a novel Incremental Boosting CNN (IB-CNN) to integrate boosting into the CNN via an incremental boosting layer that selects discriminative neurons from the lower layer and is incrementally updated on successive mini-batches. In addition, a novel loss function that accounts for errors from both the incremental boosted classifier and individual weak classifiers was proposed to fine-tune the IB-CNN. Experimental results on four benchmark AU databases have demonstrated that the IB-CNN yields significant improvement over the traditional CNN and the boosting CNN without incremental learning, as well as outperforming the state-of-the-art CNN-based methods in AU recognition. The improvement is more impressive for the AUs that have the lowest frequencies in the databases.

## 1 Introduction

Facial behavior is a powerful means to express emotions and to perceive the intentions of a human. Developed by Ekman and Friesen [1], the Facial Action Coding System (FACS) describes facial behavior as combinations of facial action units (AUs), each of which is anatomically related to the contraction of a set of facial muscles. In addition to applications in human behavior analysis, an automatic AU recognition system has great potential to advance emerging applications in human-computer interaction (HCI), such as online/remote education, interactive games, and intelligent transportation, as well as to push the frontier of research in psychology.

Recognizing facial AUs from spontaneous facial expressions is challenging because of subtle facial appearance changes, free head movements, and occlusions, as well as limited AU-coded training images. As elaborated in the survey papers [2, 3], a number of approaches have been developed to extract features from videos or static images to characterize facial appearance or geometrical changes caused by target AUs. Most of them employed hand-crafted features, which, however, are not designed and optimized for facial AU recognition. Most recently, CNNs have achieved incredible success in different applications such as object detection and categorization, video analysis, and have shown promise on facial expression and AU recognition [4, 5, 6, 7, 8, 9, 10].

CNNs contain a large number of parameters, especially as the network becomes deeper. To achieve satisfactory performance, a large number of training images are required and a mini-batch strategy is used to deal with large training data, where a small batch of images are employed in each iteration. In contrast to the millions of training images employed in object categorization and detection, AU-coded training images are limited and usually collected from a small population, e.g., 48,000 images from 15 subjects in the FERA2015 SEMAINE database [11], and 130,814 images from 27 subjects in Denver Intensity of Spontaneous Facial Action (DISFA) database [12]. As a result, the learned CNNs are often overfitted and do not generalize well to unseen subjects.

Boosting, e.g., AdaBoost, is a popular ensemble learning technique, which combines many "weak" classifiers and has been demonstrated to yield better generalization performance in AU recognition [13]. Boosting can be integrated into the CNN such that discriminative neurons are selected and activated in each iteration of CNN learning. However, the boosting CNN (B-CNN) can overfit due to the limited training data in each mini-batch. Furthermore, the information captured in previous iteration/batch cannot be propagated, i.e., a new set of weak classifiers is selected in every iteration and the weak classifiers learned previously are discarded.

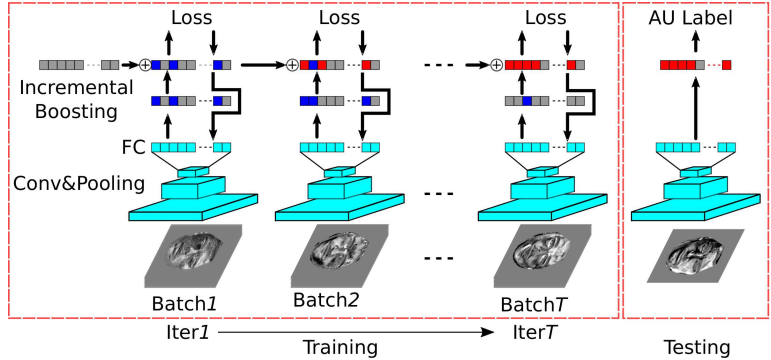

Figure 1: An overview of Incremental Boosting CNN. An incremental boosted classifier is trained iteratively. Outputs of the FC layer are employed as input features and a subset of features (the blue nodes) are selected by boosting. The selected features in the current iteration are combined with those selected previously (the red nodes) to form an incremental strong classifier. A loss is calculated based on the incremental classifier and propagated backward to fine-tune the CNN parameters. The gray nodes are inactive and thus, not selected by the incremental strong classifier. Given a testing image, features are calculated via the CNN and fed to the boosted classifier to predict the AU label. Best viewed in color.

Inspired by incremental learning, we proposed a novel Incremental Boosting CNN (IB-CNN), which aims to accumulate information in B-CNN learning when new training samples appear. As shown in Figure 1, a batch of images is employed in each iteration of CNN learning. The outputs of the fully-connected (FC) layer are employed as features; a subset of features (the blue nodes), which is discriminative for recognizing the target AU in the current batch, is selected by boosting. Then, these selected features are combined with the ones selected previously (the red nodes) to form an incremental strong classifier. The weights of active features, i.e., both the blue and the red nodes, are updated such that the features selected most of the time have higher weights. Finally, a loss, i.e., the overall classification error from both weak classifiers and the incremental strong classifier, is calculated and backpropagated to fine-tune the CNN iteratively. The proposed IB-CNN has a complex decision boundary due to boosting and is capable of alleviating the overfitting problem for the mini-batches by taking advantage of incremental learning.

In summary, this paper has three major contributions. (1) Feature selection and classification are integrated with CNN optimization in a boosting CNN framework. (2) A novel incremental boosted classifier is updated iteratively by accumulating information from multiple batches. (3) A novel loss function, which considers the overall classification error of the incremental strong classifier and individual classification errors of weak learners, is developed to fine-tune the IB-CNN.

Experimental results on four benchmark AU-coded databases, i.e., Cohn-Kanade (CK) [25] databse, FERA2015 SEMAINE database [11], FERA2015 BP4D database [11], and Denver Intensity of Spontaneous Facial Action (DISFA) database [12] have demonstrated that the proposed IB-CNN significantly outperforms the traditional CNN model as well as the state-of-the-art CNN-based methods for AU recognition. Furthermore, the performance improvement of the infrequent AUs is more impressive, which demonstrates that the proposed *IB-CNN* is capable of improving CNN learning with limited training data. In addition, the performance of IB-CNN is not sensitive to the number of neurons in the FC layer and the learning rate, which are favored traits in CNN learning.

## 2 Related Work

As detailed in the survey papers [2, 3], various human-designed features are adopted in recognizing facial expressions and AUs including Gabor Wavelets [13], Local Binary Patterns (LBP) [14], Histogram of Oriented Gradients (HOG) [15], Scale Invariant Feature Transform (SIFT) features [16], Histograms of Local Phase Quantization (LPQ) [17], and their spatiotemporal extensions [17, 18, 19]. Recently, feature learning approaches including sparse coding [20] and deep learning [4, 5, 6, 7, 8, 9, 10, 21] have been devoted to recognizing facial expressions and AUs.

Among the feature learning based methods, CNNs [4, 5, 6, 7, 8, 9, 10] have attracted increasing attention. Gudi et al. [9] used a pre-processing method with local and global contrast normalization to improve the inputs of CNNs. Fasel [4] employed multi-size convolutional filters to learn multi-scale features. Liu et al [7] extracted spatiotemporal features using the 3D CNN. Jung et al. [8] jointly fine-tuned temporal appearance and geometry features. Jaiswal and Valstar [10] integrated bi-directional long-term memory neural networks with the CNN to extract temporal features.

Most of CNN-based methods make decisions using inner product of the FC layer. A few approaches developed new objective functions to improve recognition performance. Tang [22, 6] replaced the softmax loss function with an SVM for optimization. Hinton et al. [23] utilized a dropout technique to reduce overfitting by dropping out some neuron activations from the previous layer, which can be seen as an ensemble of networks sharing the same weights. However, the dropout process is random regardless the discriminative power of individual neurons. In contrast, the proposed IB-CNN effectively selects the more discriminative neurons and drops out noisy or redundant neurons.

Medera and Babinec [24] adopted incremental learning using multiple CNNs trained individually from different subsets and additional CNNs are trained given new samples. Then, the prediction is calculated by weighted majority-voting of the outputs of all CNNs. However, each CNN may not have sufficient training data, which is especially true with limited AU-coded data. Different from [24], the IB-CNN has only one CNN trained along with an incremental strong classifier, where weak learners are updated over time by accumulating information from multiple batches. Liu et al. [21] proposed a boosted deep belief network for facial expression recognition, where each weak classifier is learned exclusively from an image patch. In contrast, weak classifiers are selected from an FC layer in the proposed IB-CNN and thus, learned from the whole face.

## 3 Methodology

As illustrated in Figure 1, an IB-CNN model is proposed to integrate boosting with the CNN at the decision layer with an incremental boosting algorithm, which selects and updates weak learners over time as well as constructs an incremental strong classifier in an online learning manner. There are three major steps for incremental boosting: selecting and activating neurons (blue nodes) from the FC layer by boosting, combining the activated neurons from different batches (blue and red nodes) to form an incremental strong classifier, and fine-tuning the IB-CNN by minimizing the proposed loss function. In the following, we start with a brief review of CNNs and then, describe the three steps of incremental boosting in detail.

### 3.1 A Brief Review of CNNs

A CNN consists of a stack of layers such as convolutional layers, pooling layers, rectification layers, FC layers, and a decision layer and transforms the input data into a highly nonlinear representation. Ideally, learned filters should activate the image patches related to the recognition task, i.e., detecting AUs in this work. Neurons in an FC layer have full connections with all activations in the previous layer. Finally, high-level reasoning is done at the decision layer, where the number of outputs is equal to the number of target classes. The score function used by the decision layer is generally the inner product of the activations in the FC layer and the corresponding weights. During CNN training, a loss layer is employed after the decision layer to specify how to penalize the deviations between the predicted and true labels, where different types of loss functions have been employed, such as softmax, SVM, and sigmoid cross entropy. In this paper, we substitute the inner-product score function with a boosting score function to achieve a complex decision boundary.

### 3.2 Boosting CNN

In a CNN, a mini-batch strategy is often used to handle large training data. Let $\mathbf{X} = [\mathbf{x}_1, ..., \mathbf{x}_M]$ be the activation features of a batch with $M$ training images, where the dimension of the activation feature vector $\mathbf{x}_i$ is $K$, and $\mathbf{y} = [y_1, ..., y_M], y_i \in \{-1, 1\}$ is a vector storing the ground truth labels. With the boosting algorithm, the prediction is calculated by a strong classifier $H(\cdot)$ that is the weighted summation of weak classifiers $h(\cdot)$ as follows:

$$H(\mathbf{x}_i) = \sum_{j=1}^{K} \alpha_j h(x_{ij}, \lambda_j); \ h(x_{ij}, \lambda_j) = \frac{f(x_{ij}, \lambda_j)}{\sqrt{f(x_{ij}, \lambda_j)^2 + \eta^2}} \quad (1)$$

where $x_{ij} \in \mathbf{x}_i$ is the $j^{th}$ activation feature of the $i^{th}$ image. Each feature corresponds to a candidate weak classifier $h(x_{ij}, \lambda_j)$ with output in the range of (-1,1). $\frac{f(\cdot)}{\sqrt{f(\cdot)^2 + \eta^2}}$ is used to simulate a $sign(\cdot)$ function to compute the derivative for gradient descent optimization. In this work, $f(x_{ij}, \lambda_j) \in \mathbb{R}$ is defined as a one-level decision tree (a decision stump) with the threshold of $\lambda_j$, which has been widely used in AdaBoost. The parameter $\eta$ in Eq. 1 is employed to control the slope of function $\frac{f(\cdot)}{\sqrt{f(\cdot)^2+\eta^2}}$ and can be set according to the distribution of $f(\cdot)$ as $\eta = \frac{\sigma}{c}$, where $\sigma$ is the standard deviation of $f(\cdot)$ and $c$ is a constant. In this work, $\eta$ is empirically set to $\frac{\sigma}{2}$. $\alpha_j \geq 0$ is the weight of the $j^{th}$ weak classifier and $\sum_{j=1}^{K} \alpha_j = 1$. When $\alpha_j = 0$, the corresponding neuron is inactive and will not go through the feedforward and backpropagation process.

Traditional boosting algorithms only consider the loss of the strong classifier, which can be dominated by some weak classifiers with large weights, potentially leading to overfitting. To account for classification errors from both the strong classifier and the individual classifiers, the loss function is defined as the summation of a strong-classifier loss and a weak-classifier loss as follows:

$$\varepsilon^B = \beta \varepsilon^B_{strong} + (1-\beta)\varepsilon_{weak} \quad (2)$$

where $\beta \in [0, 1]$ balances the strong-classifier loss and the weak-classifier loss.

The strong-classifier loss is defined as the Euclidean distance between the prediction and the groundtruth label:

$$\varepsilon^B_{strong} = \frac{1}{M} \sum_{i=1}^{M} (H(\mathbf{x}_i) - y_i)^2 \quad (3)$$

The weak-classifier loss is defined as the summation of the individual losses of all weak classifiers:

$$\varepsilon_{weak} = \frac{1}{MN} \sum_{i=1}^{M} \sum_{\substack{1 \leq j \leq K \\ \alpha_j > 0}} [h(x_{ij}, \lambda_j) - y_i]^2 \quad (4)$$

where the constraint $\alpha_j > 0$ excludes inactive neurons when calculating the loss.

Driven by the loss $\varepsilon^B$ defined in Eq. 2, the B-CNN can be iteratively fine-tuned by backpropagation as illustrated in the top of Figure 2. However, the information captured previously, e.g., the weights and thresholds of the active neurons, is discarded for a new batch. Due to limited data in each mini-batch, the trained B-CNN can be overfitted.

### 3.3 Incremental Boosting

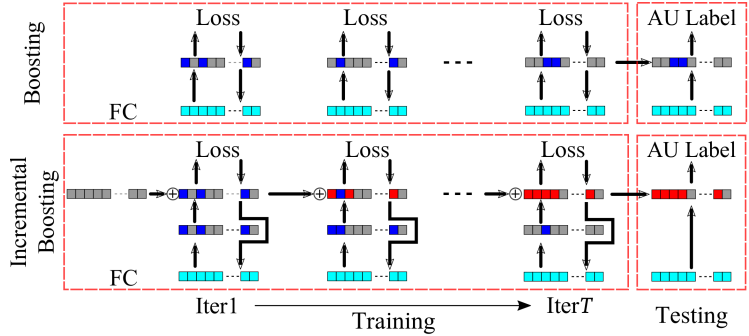

Figure 2: A comparison of the IB-CNN and the B-CNN structures. For clarity, the illustration of IB-CNN or B-CNN starts from the FC layer (the cyan nodes). The blue nodes are active nodes selected in the current iteration; the red nodes are the active nodes selected from previous iterations; and the gray nodes are inactive.

Incremental learning can help to improve the prediction performance and to reduce overfitting. As illustrated in the bottom of Figure 2, both of the blue nodes selected in the current iteration and the red nodes selected previously are incrementally combined to form an incremental strong classifier $H_I^t$ at the $t^{th}$ iteration:

$$H_I^t(\mathbf{x}_i) = \frac{(t-1)H_I^{t-1}(\mathbf{x}_i) + H^t(\mathbf{x}_i)}{t} \quad (5)$$

where $H_I^{t-1}(\mathbf{x}_i)$ is the incremental strong classifier obtained at the $(t-1)^{th}$ iteration; and $H^t(\mathbf{x}_i)$ is the boosted strong classifier estimated in the current iteration.

Substituting Eq. 1 into Eq. 5, we have

$$H_I^t(\mathbf{x}_i) = \sum_{j=1}^{K} \frac{(t-1)\alpha_j^{t-1} + \alpha_j^t}{t} h^t(x_{ij}; \lambda_j) \quad (6)$$

where $\alpha_j^t$ is calculated in the $t^{th}$ iteration by boosting. As shown in Figure 3, $h^{t-1}(\cdot)$ has been updated to $h^t(\cdot)$ by updating the threshold $\lambda_j^{t-1}$ to $\lambda_j^t$. If the $j^{th}$ weak classifier was not selected

**Algorithm 1** Incremental Boosting Algorithm for the IB-CNN

**Input:** The number of iterations (mini-batches) $T$ and activation features $\mathbf{X}$ with the size of $M \times K$, where $M$ is the number of images in a mini-batch and $K$ is the dimension of the activation feature vector for one image.
1: **for** each input activation $j$ from 1 to $K$ **do**
2: $\quad \alpha_j^1 = 0$
3: **end for**
4: **for** each mini-batch $t$ from 1 to $T$ **do**
5: $\quad$ Feed-forward to the FC layer;
6: $\quad$ Select active features by boosting and calculate weights $\boldsymbol{\alpha}^t$ based on the standard AdaBoost;
7: $\quad$ Update the incremental strong classifier as Eq. 6;
8: $\quad$ Calculate the overall loss of IB-CNN as Eq. 8;
9: $\quad$ Backpropagate the loss based on Eq. 9 and Eq. 10;
10: $\quad$ Continue backpropagation to lower layers.
11: **end for**

before, $\lambda_j^t$ is estimated in the $t^{th}$ iteration by boosting. Otherwise, $\lambda_j^t$ will be updated from the previous iteration after backpropagation as follows:

$$\lambda_j^t = \lambda_j^{t-1} - \gamma \nabla \frac{\partial \varepsilon^{H_I}}{\partial \lambda_j^{t-1}} \quad (7)$$

where $\gamma$ is the learning rate.

Then, the incremental strong classifier $H_I^t$ is updated over time. As illustrated in Figure 3, if a neuron is activated in the current iteration, the corresponding weight will increase; otherwise, it will decrease. The summation of weights of all weak classifiers will be normalized to 1. Hence, the weak classifiers selected most of the time, i.e., effective for most of mini-batches, will have higher weights. Therefore, the overall loss of IB-CNN is calculated as

Figure 3: An illustration of constructing the incremental strong classifier. Squares represent neuron activations. The gray nodes are inactive; while the blue and red nodes are active nodes selected in the current iteration and previous iterations, respectively.

$$\varepsilon^{IB} = \beta \varepsilon_{strong}^{IB} + (1-\beta) \varepsilon_{weak} \quad (8)$$

where $\varepsilon_{strong}^{IB} = \frac{1}{M} \sum_{i=1}^{M} (H_I(\mathbf{x}_i) - y_i)^2$.

Compared to the B-CNN, the IB-CNN exploits the information from all mini-batches. For testing, IB-CNN uses the incremental strong classifier, while the B-CNN employs the strong classifier learned from the last iteration.

### 3.4 IB-CNN Fine-tuning

A stochastic gradient decent method is utilized for fine-tuning the IB-CNN, i.e., updating IB-CNN parameters, by minimizing the loss in Eq. 8. The decent directions for $x_{ij}$ and $\lambda_j$ can be calculated as follows:

$$\frac{\partial \varepsilon^{IB}}{\partial x_{ij}} = \beta \frac{\partial \varepsilon_{strong}^{IB}}{\partial H_I(\mathbf{x}_i)} \frac{\partial H_I(\mathbf{x}_i)}{\partial x_{ij}} + (1-\beta) \frac{\partial \varepsilon_{weak}^{IB}}{\partial h(x_{ij};\lambda_j)} \frac{\partial h(x_{ij};\lambda_j)}{\partial x_{ij}} \quad (9)$$

$$\frac{\partial \varepsilon^{IB}}{\partial \lambda_j} = \sum_{i=1}^{M} \beta \frac{\partial \varepsilon_{strong}^{IB}}{\partial H_I(\mathbf{x}_i)} \frac{\partial H_I(\mathbf{x}_i)}{\partial \lambda_j} + (1-\beta) \sum_{i=1}^{M} \frac{\partial \varepsilon_{weak}^{IB}}{\partial h(x_{ij};\lambda_j)} \frac{\partial h(x_{ij};\lambda_j)}{\partial \lambda_j} \quad (10)$$

where $\frac{\partial \varepsilon^{IB}}{\partial x_{ij}}$ and $\frac{\partial \varepsilon^{IB}}{\partial \lambda_j}$ are only calculated for the active nodes of incremental boosting (the red and blue nodes in Figure 3). $\frac{\partial \varepsilon^{IB}}{\partial x_{ij}}$ can be further backpropagated to the lower FC layers and convolutional layers. The incremental boosting algorithm for the IB-CNN is summarized in Algorithm 1.

## 4 Experiments

To evaluate effectiveness of the proposed IB-CNN model, extensive experiments have been conducted on four benchmark AU-coded databases. The CK database [25] contains 486 image sequences from 97 subjects and has been widely used for evaluating the performance of AU recognition. In addition, 14 AUs were annotated frame-by-frame [30] for training and evaluation. The FERA2015 SEMAINE database [11] contains 6 AUs and 31 subjects with 93,000 images. The FERA2015 BP4D database [11] has 11 AUs and 41 subjects with 146,847 images. The DISFA database [12] has 12 labeled AUs and 27 subjects with 130,814 images.

### 4.1 Pre-Processing

Face alignment is conducted to reduce variation in face scale and in-plane rotation across different facial images. Specifically, the face regions are aligned based on three fiducial points: the centers of the two eyes and the mouth, and scaled to a size of $128 \times 96$. In order to alleviate face pose variations, especially out-of-plane rotations, face images are further warped to a frontal view based on landmarks that are less affected by facial expressions including landmarks along the facial contour, two eye centers, the nose tip, the mouth center, and on the forehead. A total of 23 landmarks that are less affected by facial muscle movements are selected as control points to warp the face region to the mean facial shape calculated from all images [1].

Time sequence normalization is used to reduce identity-related information and highlight appearance and geometrical changes due to activation of AUs. Particularly, each image is normalized based on the mean and the standard deviation calculated from a short video sequence containing at least 800 continuous frames at a frame rate of 30fps [2].

### 4.2 CNN Implementation Details

The proposed IB-CNN is implemented based on a modification of cifar10_quick in Caffe [28]. As illustrated in Figure 1, the preprocessed facial images are fed into the network as input. The IB-CNN consists of three stacked convolutional layers with activation functions, two maxpooling layers, an FC layer, and the proposed IB layer to predict the AU label. Specifically, the first two convolutional layers have 32 filters with a size of $5 \times 5$ and a stride of 1. Then, the output feature maps are sent to a rectified layer followed by the maxpooling layer with a downsampling stride of 3. The last convolutional layer has 64 filters with a size of $5 \times 5$, and the output $9 \times 5$ feature maps are fed into an FC layer with 128 nodes. The outputs of the FC layer are sent to the proposed IB layer. The stochastic gradient descent, with a momentum of 0.9 and a mini-batch size of 100, is used for training the CNN for each target AU.

### 4.3 Experimental Results

To demonstrate effectiveness of the proposed *IB-CNN*, two baseline methods are employed for comparison. The first method, denoted as *CNN*, is a traditional CNN model with a sigmoid cross entropy decision layer. The second method, denoted as *B-CNN*, is the boosting CNN described in Section 3.2. Both *CNN* and *B-CNN* have the same architecture as the *IB-CNN* with different decision layers.

**Performance evaluation on the SEMAINE database:** All the models compared were trained on the training set and evaluated on the validation set. The training-testing process was repeated 5 times. The mean and standard deviation of F1 score and two-alternative forced choice (2AFC) score are calculated from the 5 runs for each target AU. As shown in Table 1, the proposed *IB-CNN* outperforms the traditional *CNN* in term of the average F1 score (0.416 vs 0.347) and the average 2AFC score (0.775 vs 0.735). Not surprisingly, *IB-CNN* also beats *B-CNN* obviously: the average F1 score increases from 0.310 (*B-CNN*) to 0.416 (*IB-CNN*) and the average 2AFC score increases from 0.673 (*B-CNN*) to 0.775 (*IB-CNN*), thanks to incremental learning over time. In addition, *IB-CNN* considering both strong and weak classifier losses outperforms the one with only strong-classifier loss, denoted as *IB-CNN-S*. Note that, *IB-CNN* achieves a significant improvement for recognizing AU28 (Lips suck), which has the least number of occurrences (around 1.25% positive samples) in the training set, from 0.280 (*CNN*) and 0.144 (*B-CNN*) to 0.490 (*IB-CNN*) in terms of F1 score. The performance of *B-CNN* is the worst for infrequent AUs due to the limited positive samples in each mini-batch. In contrast, the proposed *IB-CNN* improves CNN learning significantly with limited training data.

Table 1: Performance comparison of *CNN*, *B-CNN*, *IB-CNN-S*, and *IB-CNN* on the SEMAINE database in terms of F1 and 2AFC. The format is mean±std. PPos: percentage of positive samples in the training set.

| AUs | PPos | CNN | | B-CNN | | IB-CNN-S | | IB-CNN | |
|---|---|---|---|---|---|---|---|---|---|
| | | F1 | 2AFC | F1 | 2AFC | F1 | 2AFC | F1 | 2AFC |
| AU2 | 13.5% | 0.314±0.065 | 0.715±0.076 | 0.241±0.073 | 0.646±0.060 | **0.414**±0.016 | 0.812±0.010 | 0.410±0.024 | **0.820**±0.009 |
| AU12 | 17.6% | 0.508±0.023 | 0.751±0.009 | **0.555**±0.007 | 0.746±0.013 | 0.549±0.016 | 0.773±0.007 | 0.539±0.013 | **0.777**±0.005 |
| AU17 | 1.9% | **0.288**±0.020 | 0.767±0.014 | 0.204±0.048 | 0.719±0.036 | 0.248±0.048 | 0.767±0.011 | 0.248±0.007 | **0.777**±0.012 |
| AU25 | 17.7% | 0.358±0.033 | 0.635±0.011 | **0.407**±0.006 | 0.618±0.011 | 0.378±0.009 | 0.638±0.011 | 0.401±0.014 | **0.638**±0.003 |
| AU28 | 1.25% | 0.280±0.111 | 0.840±0.076 | 0.144±0.092 | 0.639±0.195 | 0.483±0.069 | 0.898±0.006 | **0.490**±0.078 | **0.904**±0.011 |
| AU45 | 19.7% | 0.333±0.036 | 0.702±0.022 | 0.311±0.016 | 0.668±0.019 | **0.401**±0.009 | **0.738**±0.010 | 0.398±0.005 | 0.734±0.005 |
| AVG | - | 0.347±0.026 | 0.735±0.014 | 0.310±0.015 | 0.673±0.028 | 0.412±0.018 | 0.771±0.003 | **0.416**±0.018 | **0.775**±0.004 |

---

[1] For the CK, SEMAINE, and DISFA databases, 66 landmarks are detected [26] for face alignment and warping. For the BP4D database, the 49 landmarks provided in the database are used for face alignment.

[2] Psychological studies show that each AU activation ranges from 48 to 800 frames at 30fps [27].

Table 2: Performance comparison of *CNN*, *B-CNN*, and *IB-CNN* on the DISFA database in terms of F1 score and 2AFC score. The format is mean±std. PPos: percentage of positive samples in the whole database.

| AUs | PPos | CNN | | B-CNN | | IB-CNN | |
|---|---|---|---|---|---|---|---|
| | | F1 | 2AFC | F1 | 2AFC | F1 | 2AFC |
| AU1 | 6.71% | 0.257±0.200 | 0.724±0.116 | 0.259±0.150 | **0.780**±0.079 | **0.327**±0.204 | 0.773±0.119 |
| AU2 | 5.63% | 0.346±0.226 | 0.769±0.119 | 0.333±0.197 | 0.835±0.085 | **0.394**±0.219 | **0.849**±0.073 |
| AU4 | 18.8% | 0.515±0.208 | 0.820±0.116 | 0.446±0.186 | 0.793±0.083 | **0.586**±0.104 | **0.886**±0.060 |
| AU5 | 2.09% | 0.195±0.129 | 0.780±0.154 | 0.184±0.114 | 0.749±0.279 | **0.312**±0.153 | **0.887**±0.076 |
| AU6 | 14.9% | 0.619±0.072 | 0.896±0.042 | 0.596±0.086 | 0.906±0.040 | **0.624**±0.069 | **0.917**±0.026 |
| AU9 | 5.45% | 0.340±0.131 | 0.859±0.081 | 0.331±0.115 | 0.895±0.057 | **0.385**±0.137 | **0.900**±0.057 |
| AU12 | 23.5% | 0.718±0.063 | 0.943±0.028 | 0.686±0.083 | 0.913±0.030 | **0.778**±0.047 | **0.953**±0.020 |
| AU15 | 6.01% | 0.174±0.132 | 0.586±0.174 | **0.224**±0.120 | **0.753**±0.091 | 0.135±0.122 | 0.511±0.226 |
| AU17 | 9.88% | 0.281±0.154 | 0.678±0.125 | 0.330±0.132 | **0.763**±0.086 | **0.376**±0.222 | 0.742±0.148 |
| AU20 | 3.46% | 0.134±0.113 | 0.604±0.155 | **0.184**±0.101 | **0.757**±0.083 | 0.126±0.069 | 0.628±0.151 |
| AU25 | 35.2% | 0.716±0.111 | 0.890±0.064 | 0.670±0.064 | 0.844±0.049 | **0.822**±0.076 | **0.922**±0.063 |
| AU26 | 19.1% | 0.563±0.152 | 0.810±0.073 | 0.507±0.131 | 0.797±0.054 | **0.578**±0.155 | **0.876**±0.039 |
| AVG | - | 0.405±0.055 | 0.780±0.036 | 0.398±0.059 | 0.815±0.031 | **0.457**±0.067 | **0.823**±0.031 |

Table 3: Performance comparison with the state-of-the-art methods on four benchmark databases in terms of common metrics. ACC: Average classification rate.

| CK | | SEMAINE | | BP4D | | DISFA | | |
|---|---|---|---|---|---|---|---|---|
| Methods | ACC | Methods | F1 | Methods | F1 | Methods | 2AFC | ACC |
| AAM [29] | 0.955 | LGBP [11] | 0.351 | LGBP [11] | 0.580 | Gabor [12] | N/A | 0.857 |
| Gabor+DBN [30] | 0.933 | CNN [9] | 0.341 | CNN [9] | 0.522 | BGCS [31] | N/A | 0.868 |
| LBP [32] | 0.949 | DLA-SIFT [16] | 0.435 | DLA-SIFT [16] | 0.591 | LPQ [17] | 0.810 | N/A |
| | | | | | | ML-CNN [33] | 0.757 | 0.846 |
| CNN (baseline) | 0.937 | CNN (baseline) | 0.347 | CNN (baseline) | 0.510 | CNN (baseline) | 0.780 | 0.839 |
| **IB-CNN** | 0.951 | **IB-CNN** | 0.416 | **IB-CNN** | 0.578 | **IB-CNN** | 0.825 | 0.858 |

**Performance evaluation on the DISFA database:** A 9-fold cross-validation strategy is employed for the DISFA database, where 8 subsets of 24 subjects were utilized for training and the remaining one subset of 3 subjects for testing. For each fold, the training-testing process was repeated 5 times. The mean and standard deviation of the F1 score and the 2AFC score are calculated from the $5 \times 9$ runs for each target AU and reported in Table 2. As shown in Table 2, the proposed *IB-CNN* improves the performance from 0.405 (*CNN*) and 0.398 (*B-CNN*) to 0.457 (*IB-CNN*) in terms of the average F1 score and from 0.780 (*CNN*) and 0.815 (*B-CNN*) to 0.823 (*IB-CNN*) in terms of 2AFC score. Similar to the results on the SEMAINE database, the performance improvement of the infrequent AUs is more impressive. AU5 (upper lid raiser) has the least number of occurrences, i.e., 2.09% positive samples, in the DISFA database. The recognition performance increases from 0.195 (*CNN*) and 0.184 (*B-CNN*) to 0.312 (*IB-CNN*) in terms of the average F1 score.

**Comparison with the State-of-the-Art methods:** We further compare the proposed IB-CNN with the state-of-the-art methods, especially the CNN-based methods, evaluated on the four benchmark databases using the metrics that are common in those papers [3]. As shown in Tables 3, the performance of IB-CNN is comparable with the state-of-the-art methods and more importantly, outperforms the CNN-based methods.

### 4.4 Data Analysis

**Data analysis of the parameter $\eta$:** The value of $\eta$ can affect the slope of the simulated $sign(\cdot)$ function and consequently, the gradient and optimization process. When $\eta$ is smaller than 0.5, the simulation is more similar to the real $sign(\cdot)$, but the derivative is near zero for most of the input data, which can cause slow convergence or divergence. An experiment was conducted to analyze the influence of $\eta = \frac{\sigma}{c}$ in Eq. 1. Specifically, an average F1 score is calculated from all AUs in the SEMAINE database while varying the value of $c$. As illustrated in Figure 4, the recognition performance in terms of the average F1 score is robust to the choice of $\eta$ when $c$ ranges from 0.5 to 16. In our experiment, $\eta$ is set to half of the standard deviation $\frac{\sigma}{2}$, empirically.

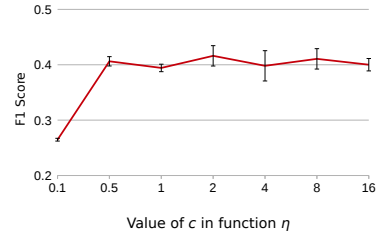

Figure 4: Recognition performance versus the choice of $\eta$.

**Data Analysis of the number of input neurons in the IB layer:** Selecting an exact number of nodes for the hidden layers remains an open question. An experiment was conducted to demonstrate that the proposed IB-CNN is insensitive to the number of input neurons. Specifically, a set of IB-

---

CNNs, with the number of input neurons of 64, 128, 256, 512, 1042, and 2048, were trained and tested on the SEMAINE database. For each IB-CNN, the average F1 score is computed over 5 runs for each AU. As shown in Figure 5, the *B-CNN* and especially, the proposed IB-CNN is more robust to the number of input neurons compared to the traditional *CNN* since a small set of neurons are active in contrast to the FC layer in the traditional CNN.

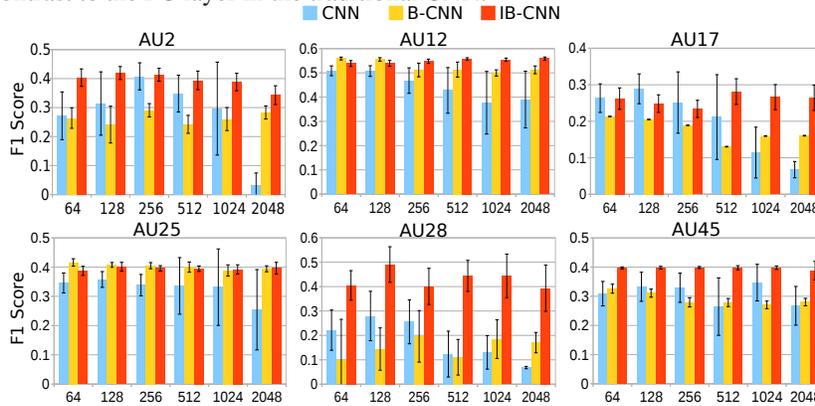

Figure 5: Recognition performance versus the number of input neurons in the IB layer.

**Data analysis of learning rate $\gamma$:** Another issue in CNNs is the choice of the learning rate $\gamma$. The performance of the IB-CNN at different learning rates is depicted in Figure 6 in terms of the average F1 score calculated from all AUs on the SEMAINE database. Compared to the traditional *CNN*, the proposed IB-CNN is less sensitive to the value of the learning rate.

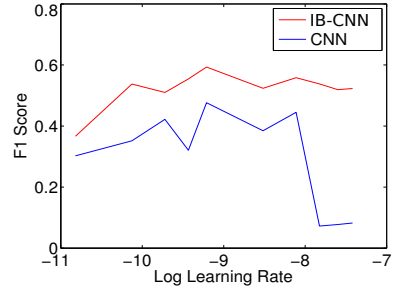

Figure 6: Recognition performance versus the learning rate $\gamma$.

## 5   Conclusion and Future Work

In this paper, a novel IB-CNN was proposed to integrate boosting classification into a CNN for the application of AU recognition. To deal with limited positive samples in a mini-batch, an incremental boosting algorithm was developed to accumulate information from multiple batches over time. A novel loss function that accounts for errors from both the incremental strong classifier and individual weak classifiers is proposed to fine-tune the IB-CNN. Experimental results on four benchmark AU databases have demonstrated that the IB-CNN achieves significant improvement over the traditional CNN, as well as the state-of-the-art CNN-based methods for AU recognition. Furthermore, the IB-CNN is more effective in recognizing infrequent AUs with limited training data. The IB-CNN is a general machine learning method and can be adapted to other learning tasks, especially those with limited training data. In the future, we plan to extend it to multitask learning by replacing the binary classifier with a multiclass boosting classifier.

## Acknowledgment

This work is supported by National Science Foundation under CAREER Award IIS-1149787.